\documentclass[final]{cvpr}

\usepackage{times}
\usepackage{epsfig}
\usepackage{graphicx}
\usepackage{amsmath}
\usepackage{amssymb}

\usepackage[breaklinks=true,bookmarks=false]{hyperref}

\usepackage{cuted}
\usepackage{capt-of}
\usepackage{placeins}

\def\cvprPaperID{****} 

\title{\LaTeX\ Author Guidelines for CVPR Proceedings}

\author{First Author\\
Institution1\\
Institution1 address\\
{\tt\small firstauthor@i1.org}
\and
Second Author\\
Institution2\\
First line of institution2 address\\
{\tt\small secondauthor@i2.org}
}

\long\def\ignorethis#1{}

\usepackage{amsmath}
\usepackage{multirow}
\usepackage{siunitx}
\usepackage{url}

\usepackage{xfrac}
\usepackage{newtxmath}
\usepackage{amsfonts,amssymb,mathrsfs}

\usepackage{calrsfs}
\DeclareMathAlphabet{\pazocal}{OMS}{zplm}{m}{n}

\usepackage{booktabs}

\usepackage{wrapfig}
\usepackage[strict]{changepage}

\usepackage{soul}
\soulregister\ref{7}
\soulregister\cite{7}
\soulregister\refFig{7}

\usepackage{color}
\definecolor{gray}{rgb}{0.35,0.35,0.35}
\definecolor{blue}{rgb}{0,0,1}
\definecolor{white}{rgb}{1,1,1}

\usepackage{booktabs}

\usepackage{epsfig}
\usepackage{epstopdf}
\usepackage{subfig}
\usepackage{multirow}
\usepackage{array}

\usepackage{xcolor}

\newbox\jsavebox

\hypersetup{
    colorlinks,
    citecolor=green
}
\begin{document}

\title{Single Pair Joint Cross-Modality Super Resolution}

\author{Guy Shacht
\qquad
Sharon Fogel
\qquad
Dov Danon
\qquad
Daniel Cohen-Or
\qquad
Ilya Leizerson
\\\\
\hspace*{4.35cm} Tel-Aviv University
\qquad
\hspace*{4.05cm} Elbit Systems
}

\maketitle

\begin{abstract}
   Non-visual imaging sensors are widely used in the industry for different purposes.
Those sensors are more expensive than visual (RGB) sensors, and usually produce images with lower resolution.
To this end, Cross-Modality Super-Resolution methods were introduced, where an RGB image of a high-resolution assists in increasing the resolution of a low-resolution modality. However, fusing images from different modalities is not a trivial task, since each multi-modal pair  varies greatly in its internal correlations. For this reason, traditional state-of-the-arts which are trained on external datasets often struggle with yielding an artifact-free result that is still loyal to the target modality characteristics.

We present CMSR, a single-pair approach for Cross-Modality Super-Resolution.
The network is internally trained on the two input images only, in a \textit{self-supervised} manner, learns their internal statistics and correlations, and applies them to up-sample the target modality. 
CMSR contains an internal transformer which is trained on-the-fly together with the up-sampling process itself and without supervision, to allow dealing with pairs that are only weakly aligned.
We show that CMSR produces state-of-the-art super resolved images, yet without introducing artifacts or irrelevant details that originate from the RGB image only.

\end{abstract}

\section{Introduction}
    Super-Resolution (SR) methods are used to increase the spatial resolution and improve the level-of-detail of digital images, while preserving the image content.  
Such methods have important applications for multiple industries, such as health-care, agriculture, defense  and film. (\cite{Nasrollahi:2014:SCS:2647753.2647819})
In recent years, more advanced methods of SR have been heavily based on Deep Learning. (\cite{DBLP:journals/corr/DongLHT15, DBLP:journals/corr/LedigTHCATTWS16, DBLP:journals/corr/abs-1904-07523})

\par
The need for super-resolution becomes even more prominent when dealing with sensors that capture other modalities, different than the visible light spectrum, since those sensors typically produce images with substantially lower resolution. (\cite{kiran2017single,Mandanici2019})
For example, Infra-Red (IR) camera sensors are more expensive than classical camera sensors, and their output images commonly have much lower spatial resolution.
%
To bridge that gap in level-of-detail, Joint Cross-Modality methods were developed. The idea is to use the higher-resolution RGB modality to guide the process of super-resolution on images taken by the lower resolution sensor, taking advantage of the finer details found in the RGB images.
The challenge is to remain loyal to the target modality characteristics and to avoid adding redundant artifacts or textures that may be present only in the RGB modality.

\par
In this work, learning is performed internally, relying solely on the input pair of images. This approach does not require any training data, and therefore avoids the need for a modal-specific dataset, relying solely on the internal image-specific statistics instead. (\cite{DBLP:journals/corr/abs-1712-06087}) Using an internal super-resolution method is \textit{particularly} strong in the context of cross-modality, since it allows the network to fit to the unique properties and the modality characteristics of the specific input pair. This feature stands in contrast to the somewhat impractical task of generalizing to a large cross-modal image dataset; each multi-modal pair is inherently unique in its internal correlations, and therefore must be treated differently. 

\par
\begin{figure*}
\centering
\hspace{-20pt} Pixel-to-Pixel SR \cite{DBLP:journals/corr/abs-1904-01501} \hspace{25pt} DeepJF \cite{DBLP:journals/corr/abs-1710-04200} \hspace{50pt} CMSR \hspace{75pt} GT \hspace{70pt} RGB Input 
\includegraphics[height=85px,width=\linewidth]{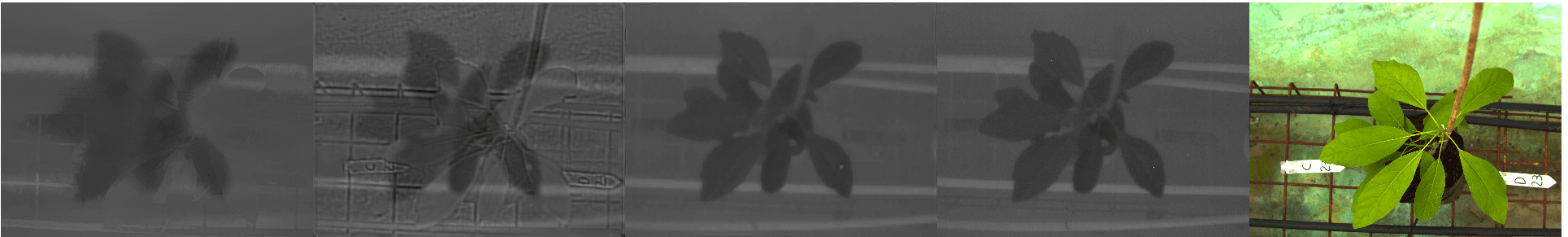}
\caption{
In this \textbf{visual-thermal} pair, the two inputs are displaced (visualised in Figure \ref{fig:deformation_vase}), a phenomenon which often occurs in real-world scenarios. This is a failure case even for state-of-the-arts, as they are based on perfect alignment. CMSR adapts to the given misalignment, and corrects it as a part of the SR process, producing a sharp result (41.069 dB PSNR). The deformed RGB image is presented in Figure \ref{fig:deformation_vase} 
}
\label{fig:maagad_deformation_comparison}
\end{figure*}

State-of-the-art Joint Cross-Modality SR methods rely on the assumption that their multiple inputs are well aligned. (\cite{almasri2018multimodal, almasri2018rgb, DBLP:journals/corr/abs-1708-09105, chen2016color, ni2017color, DBLP:journals/corr/abs-1904-01501, DBLP:journals/corr/abs-1710-04200})
Thus, they perform well only when the input images were captured by different sensors placed in the exact same position, and taken at the exact same time. As to be shown, in real-life scenarios perfect alignment of multiple sensors is often hard to achieve. 
%
In our work, we present new means to allow the two modalities to be moderately misaligned, namely weakly aligned. Our network contains a learnable deformation component that implicitly aligns details in the two images together.
More specifically, our architecture includes a deformation model that aligns details from the RGB image to the target modality in a coarse-to-fine manner, before they are fused together. The network does not use any explicit supervision for the deformation sub-task, but rather optimizes the deformation parameters to adhere to the super-resolution goal. Figures \ref{fig:maagad_deformation_comparison} and \ref{fig:vase_comparison} present cases where a weakly aligned pair causes state-of-the-arts methods to fail, whereas our method produces high-quality super-resolved output.

\par
Another notable advantage of our single-pair approach is the avoidance of over-transferal of information. 
Previous approaches which train on external cross-modal datasets are often limited in their ability to adjust to the specific nature of the input pair. For this reason, they often struggle in cases where the guiding modality should be only minimally used, or even completely ignored; they tend to fuse redundant details from the guiding modality anyway, resulting in the addition of textures and artifacts to the lower resolution modality.
Our method is designed to adjust to the specific input pair and therefore transfers details from the higher resolution image carefully and conservatively, learning only the details which aid improving the super-resolution task. Figure \ref{fig:artifacts_rgbnir} presents an example with cross-modality ambiguity. Namely, the RGB modality contains an object which does not exist in the target modality; this object should ideally be ignored. Our network successfully avoids transferring it, whereas a competing cross-modality method results in unwanted artifacts and textures.
We show that our network achieves state-of-the-art results on different modalities (Thermal, NIR, Depth), while being generic in supporting any modality as input and requiring no pre-training (and thus, no training data).


%

\section{Related Works}
    Super-Resolution has been extensively studied throughout the last two decades. 
See \cite{Nasrollahi:2014:SCS:2647753.2647819} for a survey covering various SR techniques. 
Recent surveys (\cite{DBLP:journals/corr/abs-1808-03344,DBLP:journals/corr/abs-1904-07523}) cover more advanced methods, including Deep-Learning based methods.
The first notable deep network-based method of SR method is SRCNN, (\cite{DBLP:journals/corr/DongLHT15}) a simple fully convolutional method that showed superior results to traditional methods.
Like most methods, SRCNN uses external image datasets, like T91, Set5 and Set14 (\cite{MSLapSRN,DBLP:journals/corr/LedigTHCATTWS16}) for training and evaluation.

However, it was claimed (\cite{irani2009super,zontak2011internal, DBLP:journals/corr/abs-1712-06087}) that methods which rely on large external datasets do not learn the internal image-specific properties of the given input. In \cite{irani2009super,zontak2011internal}, the subject of internal patch recurrence is investigated, and the benefits of a single-image approach were shown.
This strong observation gave rise to powerful Zero-Shot methods, (\cite{Huang_2015_CVPR,DBLP:journals/corr/abs-1712-06087,cui2014deep}) and most notably ZSSR. (\cite{DBLP:journals/corr/abs-1712-06087}) Our work extends the concept of Zero-Shot to cross-modality. This way, we not only enjoy the advantage of being dataset independent and adjusting to any modality, but we also enable our network to adapt to the specific properties (which are to be discussed) of the specific input pair.



\subsection{Joint Cross-Modality}
In the Joint Cross-Modality setting the two different modalities are jointly analyzed to enhance one of them. As mentioned earlier, camera sensors capturing the RGB modality produce images with richer HR details than other modalities. Thus, a common setting is the usage of a visual HR version of the image, alongside with a LR version taken by the other modality sensor. This setting was adopted by all relevant joint cross-modality methods.

In \cite{ni2017color}, a learning-based visual-depth method is presented. It is based on a CNN architecture operating on a LR depth-map and a sharp edge-map extracted from the HR visual modality. The network is trained on visual-depth aligned pairs from the Middlebury dataset. (\cite{Scharstein:2002:TED:598429.598475})
The method Deep Joint Image Filtering (\cite{DBLP:journals/corr/abs-1710-04200}) presented a framework for denoising and upsampling depth-maps. Their network performs concatenation of features extracted from the guiding image and features extracted from the target modality image. It was evaluated on the Middlebury dataset and has shown promising results for its task. It is however designed to be pretrained on a full multi-modal datasets of perfectly aligned pairs. Almastri et al. (\cite{almasri2018multimodal}) introduced the learning-based visual-thermal SR methods VTSRCNN and VTSRGAN, built on top of the existing SRCNN and SRGAN. They perform joint visual-thermal SR by concatenating feature maps extracted from each input modality, and are trained and evaluated on the ULB17-VT (\cite{Almasri2019}) visual-thermal dataset consisting of well aligned pairs.
In Guided Super-Resolution as Pixel-to-Pixel Transformation (\cite{DBLP:journals/corr/abs-1904-01501}), the problem of guided depth-maps SR was posed as a pixel-to-pixel translation of the HR guiding modality to a newly predicted HR depth-map, constrained by the intensities of the matching regions from the LR depth-map input. This method has shown to produce sharp HR depth-maps, but is based on perfect alignment between the input and the guiding modality.


\paragraph{\textbf{Cross-Modal Misalignment}}
In the context of cross-modality super-resolution, misalignment is a major limitation in producing artifact-free SR results. This was previously discussed in related works, (\cite{almasri2018multimodal, DBLP:journals/corr/abs-1710-04200}) and shown explicitly in this paper.
Our method's approach in handling cross-modal misalignment is to deform the RGB modality and align details that improve the SR objective to the target modality.

\paragraph{\textbf{Our Method}}
Our method differs from the aforementioned joint cross-modality techniques in two central aspects. 
First, it does not require any training data, and therefore avoids the need for a modal-specific dataset, relying on the internal image-specific statistics instead. This feature is especially attractive for cross-modality super-resolution; learning from the single input pair encourages the network to adapt to the specific cross-modal properties existing in that particular pair, which may be unique. This is unlike previous state-of-the-arts who are trained on external datasets and are often limited in their capability to adapt to a specific pair, and therefore result in over-transferal of information (such as in Figures \ref{fig:artifacts_rgbnir} and \ref{fig:artifact_comparison}).
Second, it requires only \textit{weak} alignment, as opposed to the aforementioned techniques which rely on well aligned pairs. This attribute is critical when operating in real-life scenarios. For example, state-of-the-arts like Pixel-to-Pixel SR (\cite{DBLP:journals/corr/abs-1904-01501}) and DeepJF (\cite{DBLP:journals/corr/abs-1710-04200}) struggle when applied to pairs that were captured in less-than-optimal imaging conditions, as shown most notably in Figures \ref{fig:maagad_deformation_comparison} and \ref{fig:vase_comparison}.

\subsection{Multi-Modal Alignment}
The subject of multi-modal image registration has been studied mainly in the context of medical imaging. Deep methods (\cite{DBLP:journals/corr/SimonovskyGMNK16, DBLP:journals/corr/VosBVSI17, DBLP:journals/corr/abs-1809-06130}) have mostly based their architectures on a regressor, a spatial transformer and a re-sampler. They use supervision to optimize their regression and deformation models. It is also possible to use similarity metrics like cross-correlation (\cite{DBLP:journals/corr/abs-1809-06130}) instead, and obtain an unsupervised registration framework. In \cite{DBLP:journals/corr/abs-2003-08073}, unsupervised registration was performed through an image-to-image translation objective. Namely, the better the network translates one modality to the second modality (which is the modality being deformed), the better the deformation is assumed to be done.
\par

In our work, multi-modal image registration is \textbf{not} performed per se. Our goal is not to register the two input modalities together completely, but to give the network enough freedom to align only the details that assist and adhere to the super-resolution task. For this reason, it is  possible that the network chooses to only partly align the guiding modality. The alignment phase is integrated into the main SR task. We use the same SR reconstruction loss to optimize our deformation parameters. Thus, we do not require aligned pairs for training. The deformation framework used in our method consists of three steps performed in a coarse-to-fine manner, with the help of affine, CPAB and TPS layers. (\cite{freifeld2017, skafte2018deep}) More specific details are found in later chapters. 
    
\section{Cross-Modality Super Resolution}
    One of the fundamental problems of cross-modality super resolution is that it is hard to transfer only the relevant details from the higher resolution image to the lower resolution one while ignoring unnecessary details, which often cause ghosts and unwanted artifacts (such as those in Figure \ref{fig:artifacts_rgbnir} and Figure \ref{fig:artifact_comparison}). When training on a large dataset of cross-modality image pairs, it is hard (and often impossible) for a network to learn which details exactly should be transferred and which should not be, for each given cross-modality pair. This is mostly because similar objects might be present (and thus, should be transferred) in some pairs in the dataset, and not in others. This way, externally trained networks have to decide in inference whether to incorporate some given RGB detail or not.
To avoid this problem, we opted to use a super resolution method trained on a \textit{single} input pair, and enable the network to adapt to it specifically.



\begin{figure}
\includegraphics[width=\linewidth]{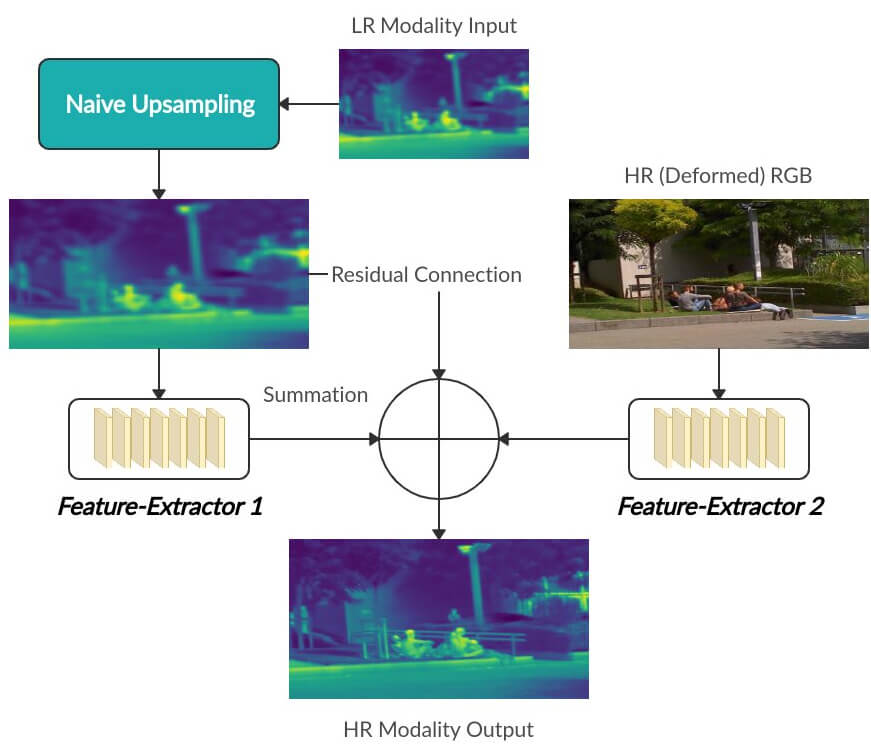}
\caption{
CMSR performs three-way summation; two of the resulting feature maps, one from each modality, are summed together with the original modality input that is naively up-sampled, in a residual manner.
}
\label{fig:architecture}
\end{figure}

\begin{figure*}
\includegraphics[width=\linewidth]{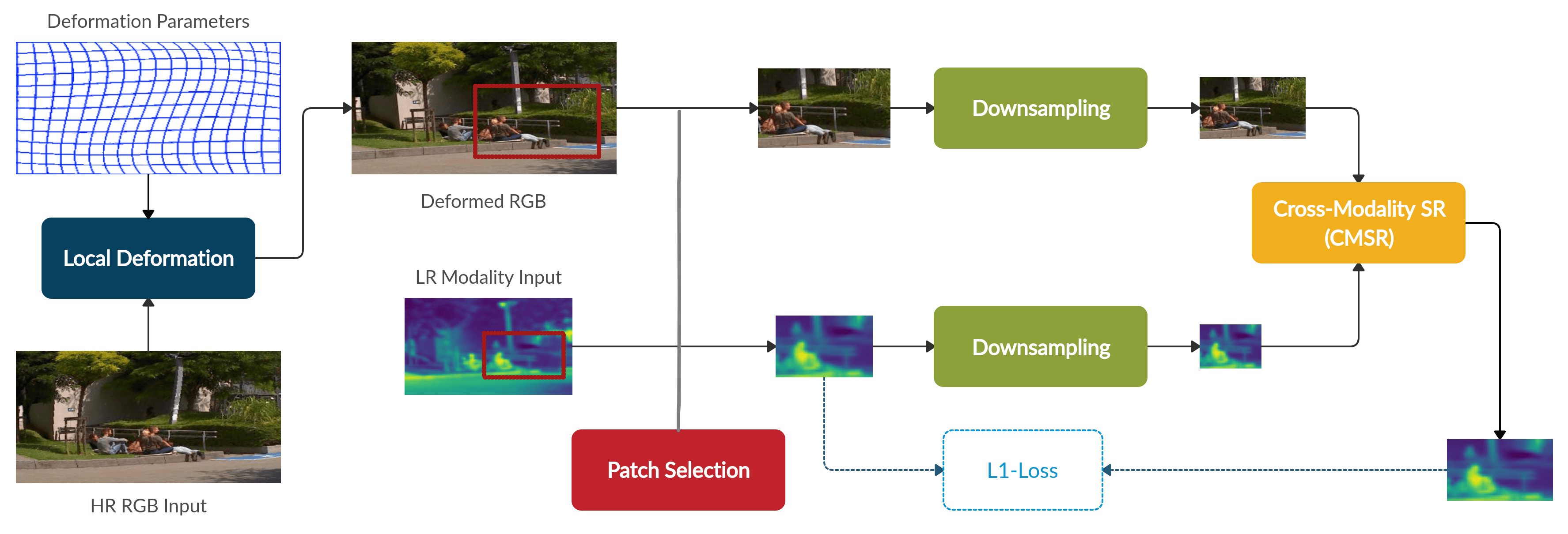}
\caption{
{\em Training process.} 
The RGB image first goes through a deformation step which aligns it to the target modality (in blue).
Then, random patches are selected by an augmentation step (in Red) and down-sampled (in green).
The patches are used to train the CMSR network (in orange) and the deformation parameters. The loss function is measured between the super-resolved output and the input target modality images.  
}
\label{fig:training}
\end{figure*}

\subsection{Network Architecture}
\label{architecture_section}
\begin{figure}
\includegraphics[width=\linewidth]{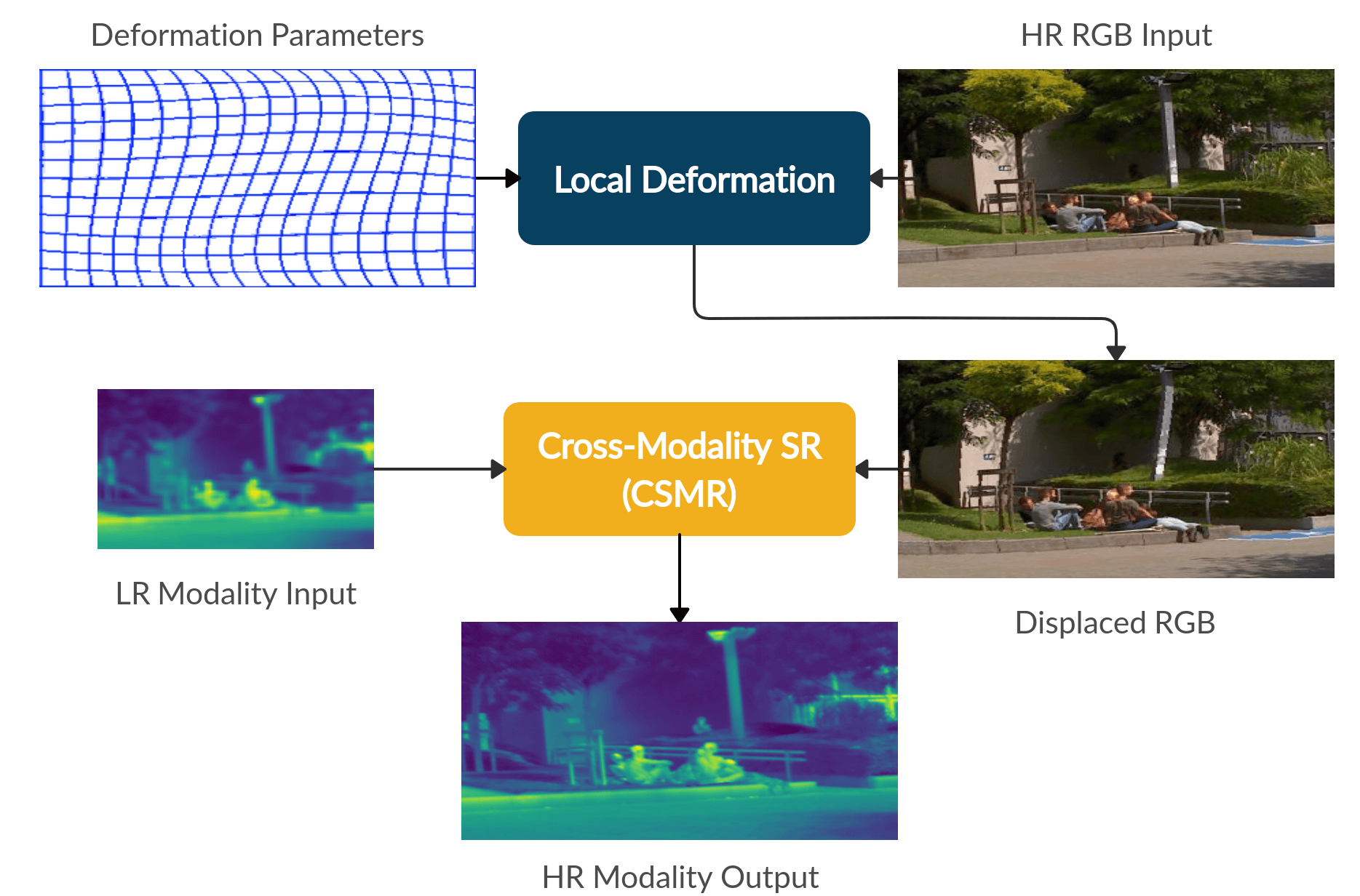}
\caption{
{\em Inference.}
During inference, the learned deformation parameters and the CMSR component are used to up-sample the original LR modality input image, guided by the HR RGB input image. 
}
\label{fig:testing}
\end{figure}

Our network includes a patch selection component which generates a training set out of a single pair of images, and a super-resolution network. Our method enables dealing with misaligned pairs by including a deformation phase, done internally, which aligns objects in both images right before they enter the SR network (see Figures \ref{fig:training} and \ref{fig:testing}). 
We hereby introduce and describe the components of our network, which are incorporated into our training and inference schemes as covered in Sections \ref{training} and \ref{inference}.

\paragraph{\textbf{Alignment using Learnable Deformation}}
\label{para:alignment}


Our network corrects displacements between the two modalities on-the-fly, through a local deformation process applied to the RGB modality as a first gate to the network, optimized implicitly during training.
To that end, instead of using explicit supervision to optimize the deformation parameters, they are trained with the super-resolution loss and therefore deform only parts which are relevant to this task. Hence, the goal of the deformation step is not to form a perfect alignment between the images, but rather to allow partial alignment to boost the super-resolution task, where needed.
Our deformation process consists of three different transformation layers, performing the learned alignment in a coarse-to-fine manner. 
\bigbreak
The first layer of our deformation framework is the original \textbf{Affine STN} layer by Jaderberg et al. (\cite{DBLP:journals/corr/JaderbergSZK15}) It captures a global affine transformation that is used to position the two modalities together as a rough initial approximation. 
\par
The second layer is a DDTN transformation layer (Deep Diffeomorphic Transformation Network, \cite{skafte2018deep}), a variant of the original STN layer supporting more flexible and expressive transformations. Our chosen transformation model is \textbf{CPAB} (Continuous
Piecewise-Affine Based, \cite{freifeld2017, skafte2018deep}). It is based on the integration of Continuous Piecewise-Affine (CPA) velocity fields, and yields a transformation that is both differentiable and has a differentiable inverse. It is Continuous Piecewise-Affine w.r.t a tessellation of the image into cells. For this reason, it is well suited to our alignment task; each cell can be deformed differently, yet continuity is preserved between neighboring cells, yielding a deformation that can express local (per-cell) misalignments while preserving the image semantics.
\par
The third and last layer of our deformation framework performs a \textbf{TPS} (Thin-plate spline) transformation, a technique that is widely used in computer vision and particularly in image registration tasks. (\cite{DBLP:journals/pami/Bookstein89})
Our implementation (also taken from \cite{skafte2018deep}) learns the displacements of uniformly-distributed keypoints in an arbitrary way, while each keypoint's surrounding pixels are displaced in accordance to it, using interpolation. (\cite{DBLP:journals/pami/Bookstein89})
Since TPS displaces its keypoints freely, the displacement is unconstrained to any image transformation model, and has the power to align the fine-grained objects of the scene, providing the final refinement of our alignment task.

\paragraph{\textbf{Patch Selection}}
\label{para:patch_selection}
We produce our training set from a single pair of images by sampling patches using random augmentations.
In our implementation we use scale, rotation, shear and translations. This random patch selection yields two patches that correspond to roughly the same area in the scene: one taken from the target modality and the second is taken from the deformed RGB modality which was previously aligned to the target modality.


\paragraph{\textbf{CMSR network}}
The CMSR network is the main component of our architecture, responsible for performing super-resolution. It produces a HR version of its target modality LR input image, guided by its HR RGB input. As Figures 
\ref{fig:training} and \ref{fig:testing} suggest, CMSR can be applied to varying image sizes, thanks to its fully convolutional nature.

The first gate to the network is up-sampling of the LR modality input to the size of the RGB input. This is done naively, using the Bi-cubic method, in case no specific kernels are given. \footnote{\label{ftnote}Optimal blur kernels can be directly estimated as shown in \cite{irani2009super}, and are fully supported by our method as an additional input to the network.}
From the up-sampled modality input we generate a feature map using a number of convolutional layers, denoted as \textbf{\textit{Feature-Extractor 1}} in Figure \ref{fig:architecture}.
From the RGB modality input that was previously aligned to target modality input, we generate a feature map using \textbf{\textit{Feature-Extractor 2.}}
We perform summation of the two resulting feature maps, one from each Feature-Extractor block, alongside with an up-sampled version of the LR target modality image, in a residual manner. 
This yields our HR super-resolved output.

\subsection{Training}
\label{training}


During each training iteration, we perform local deformation on the RGB modality input and produce a displaced version of it, aligned to the target modality image, as described in \ref{para:alignment}. 
Then, a random patch is selected from the input pair (illustrated in Figure \ref{fig:training}), yielding two corresponding patches; one taken from the target modality, and the second from the displaced (aligned) RGB modality, as described in \ref{para:patch_selection}. The patch selection phase is an integral part of the network, and is done in a differentiable manner, so as to allow the gradients to backpropagate through it to the deformation model. This enables us to optimize the transformation on the entire RGB image despite using patches of the image during training.

In order to generate supervision for the training process, we down-sample the two patches and use the original target modality patch as ground-truth.
We use $L_1$ reconstruction loss between the reconstructed patch and original input target modality patch.
Note that there is no ground truth for a perfectly aligned RGB modality. Instead, the deformation parameters are optimized using the same $L_1$ reconstruction loss as an integral part of the SR task.




\subsection{Inference}
\label{inference}
At inference time, we use the trained CMSR network and deformation parameters, to perform SR on the entire target modality image guided by the RGB modality image (see Figure \ref{fig:testing}).

Since CMSR is fully convolutional, it can operate on any image size (e.g., both image patches of different scales, and full images) using the same network.
We first apply the alignment dictated by the optimized deformation parameters, and then feed the LR target modality image and the aligned HR RGB image to the SR network which outputs a HR version of the target modality image.

After the HR target modality image is obtained, we perform two additional refinement operators aimed to further improve our SR results. The first operator, \textbf{Geometric Self-Ensemble}, is an averaging technique shown to improve SR results. (\cite{DBLP:journals/corr/LimSKNL17,DBLP:journals/corr/TimofteRG15,DBLP:journals/corr/abs-1712-06087}) The second operator, \textbf{Iterative Back-Projection}, is an error-correcting technique that was used successfully in the context of SR. (\cite{Glasner2009, Irani:1991:IRI:108693.108696,DBLP:journals/corr/abs-1712-06087})



\section{Results and Evaluation}
    \begin{figure*}
\includegraphics[height=90pt,width=\linewidth]{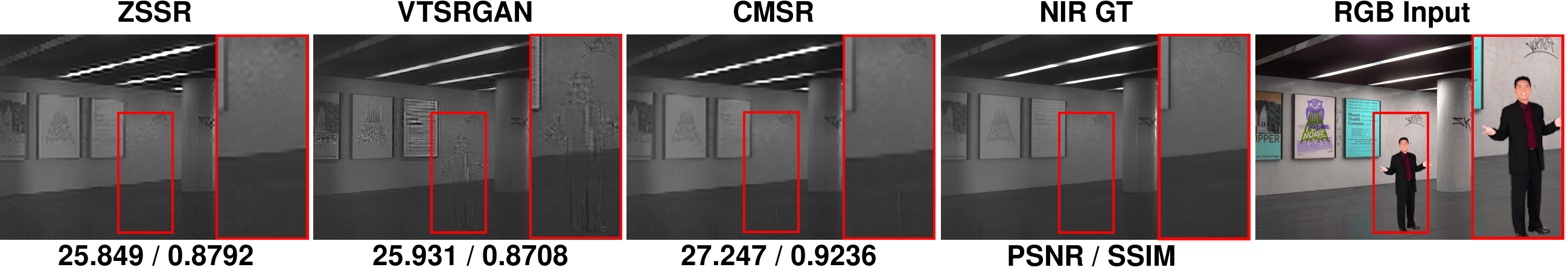}
\caption{
CMSR successfully ignores details that appear only in the RGB image (a standing man). It does not add ghosts, while VTSRGAN does. CMSR also surpasses the baseline single-modality method, ZSSR, which only operates on the NIR input.}

\label{fig:artifacts_rgbnir}
\end{figure*}

\begin{figure}
\includegraphics[width=\linewidth]{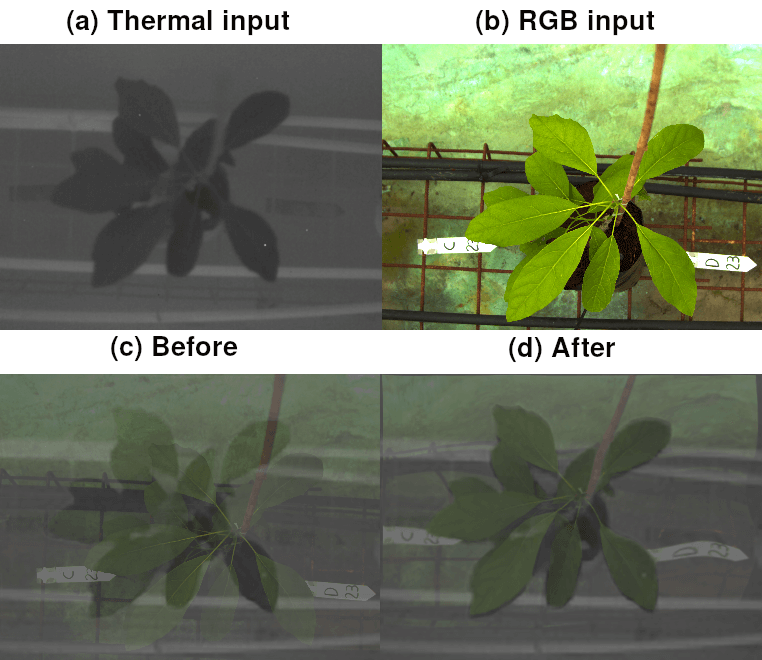}
\caption{
We evaluated CMSR on a severely misaligned visual-thermal pair, (a) and (b), with both global and local displacements.
We overlaid the images, 
once before training (c), and once after training the network (d).
CMSR deformed its RGB input on-the-fly to better alignment, relying solely on our SR loss.}
\label{fig:deformation_vase}
\end{figure}

\subsection{Implementation Details}
\label{para:implementation_details}
\label{sec:imp}
Our model is implemented in Tensorflow 1.11.0 and trained on a single GeForce GTX 1080 Ti GPU. The full code and datasets will are published in the project's GitHub page and will be deanonymized upon acceptance. 
We typically start with a learning rate of 0.0001 and gradually decrease it to $10^{-6}$, depending on the slope of our reconstruction error line, whereas the learning rates of our transformation layers follow the same pattern, multiplied by constant factors. Those factors are treated as hyper-parameters, and should typically be larger when dealing with highly displaced input pairs, like in the case of weakly aligned modalities.
Performing a $4x$ SR on an input of size $60 x 80$ typically takes around 30 seconds on a single GPU. We stop training when the reconstruction error slope does not change dramatically over a fixed number of iterations.
To achieve SR of higher scales, we perform gradual SR with intermediate scales, as this further improves the results. (\cite{DBLP:journals/corr/LaiHA017,DBLP:journals/corr/abs-1804-02900,DBLP:journals/corr/abs-1712-06087})

For \textbf{\textit{Feature-Extractor 1}} we use eight hidden layers, each containing 64 channels and a filter size of $3x3$. We place a ReLU activation function after each layer except for the last one. The size of feature maps remains the same throughout all layers in the block.
For \textbf{\textit{Feature-Extractor 2}} we typically use four to eight hidden layers with number of channels ranging from 4 to 128, a filter size of $3x3$ and a ReLU activation function. The last layer has no activation and a filter size of $1x1$. We find that highly detailed RGB inputs require \textbf{\textit{Feature-Extractor 2}} to have more channels. The hyper-parameters rarely require adjustments; they only require manual tuning when dealing with inputs that are unique, unusual, or ones that reflect very unusual displacements.

\subsection{Evaluation with State-of-the-arts}
\label{sec:exp}

\begin{figure*}
\centering
\includegraphics[width=\linewidth]{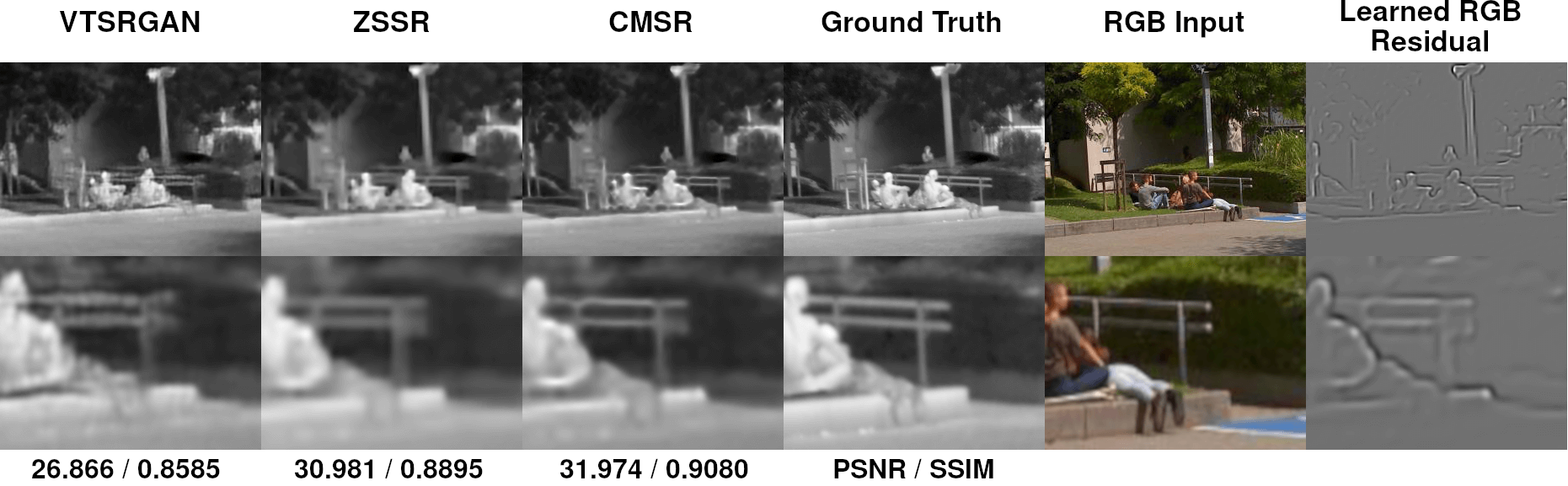}

\caption{We compare our method to its baseline method, ZSSR (\cite{ DBLP:journals/corr/abs-1712-06087}), as well as to another cross-modality method, VTSRGAN (\cite{almasri2018multimodal}) on a \textbf{visual-thermal} pair from the ULB17-VT evaluation.  On the right, the output of \textit{\textbf{Feature-Extractor 2}} (Figure \ref{fig:architecture}) is given as the learned RGB residual which is added to our output. This RGB residual resembles an edge-map; it is artifact-free and contains no unwanted textures}
\label{fig:fence_comparison}
\end{figure*}

\subparagraph{\textbf{Thermal (Infrared).}}
We compared our method to cross-modal state-of-the-art SR methods on visual-thermal pairs. We used the ULB17-VT dataset (\cite{Almasri2019}), consisting of pairs (two examples are shown in the supplementary material, Figure 5, bottom row) that are mostly well aligned.
We have included the results of our evaluation in Table \ref{table:avg_table1}, showing that our method, despite not being previously trained, beats competing methods. In Figure \ref{fig:fence_comparison} a visual result from that evaluation is included. In Figure \ref{fig:maagad_deformation_comparison} another visual-thermal example is given, taken from a visual-thermal agricultural dataset.

\subparagraph{\textbf{NIR (Near-infrared).}}

In Figures \ref{fig:artifacts_rgbnir}, \ref{fig:artifact_comparison} and in supplementary material Figure 1 (containing a more extensive evaluation), we include visual results from our evaluation on the EPFL NIR dataset. (\cite{ivrl}) The conservative approach of our method enables it to surpass state-of-the-art methods (Table \ref{table:avg_table1}), even though the competing methods were pre-trained, whereas our method operates on a single input pair without pretraining.

\begin{figure*}
\centering
\hspace{-20pt} Pixel-to-Pixel SR \cite{DBLP:journals/corr/abs-1904-01501} \hspace{30pt} DeepJF \cite{DBLP:journals/corr/abs-1710-04200} \hspace{55pt} CMSR  \hspace{70pt} GT \hspace{70pt} RGB Input 
\includegraphics[width=\linewidth]{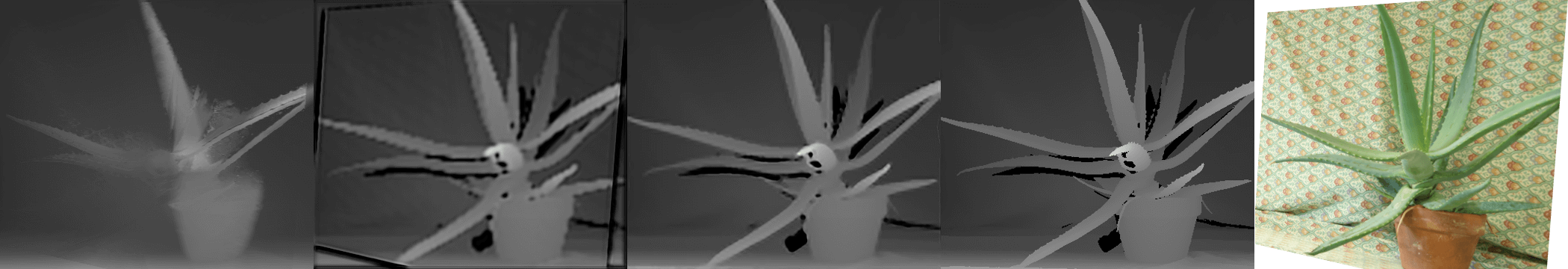}
\caption{This misaligned \textbf{visual-depth} pair is taken from our \textit{Shuffled}-Middlebury dataset evaluation. Compared to cross-modality state-of-the-arts, who struggle to produce a clear result, our method succeeds in this task (32.239 dB PSNR / 0.9403 SSIM) thanks to its alignment capabilities.}
\label{fig:vase_comparison}
\end{figure*}

\subparagraph{\textbf{Depth.}}
The Middlebury dataset (\cite{Scharstein:2002:TED:598429.598475}) contains strongly aligned depth-visual pairs as shown in the supplementary Figure 5 (top row). Multiple angles from different sensor placements are included. To obtain weakly-aligned pairs, we shuffled the pairs together such that the resulting pairs would correspond to sensor misplacements. An example is given in the supplementary material. We denote the new resulting dataset as \textit{Shuffled}-Middlebury.
CMSR surpasses competing cross-modal methods on those weakly aligned pairs by using a coarse-to-fine alignment approach, as summarized in Table \ref{table:avg_table1} and presented in Figure \ref{fig:vase_comparison}.

\begin{table*}
\centering
\begin{tabular}{|c|c|c|c|c|c|c|}
        \hline
            Metric & Dataset & ZSSR & VTSRGAN  & VTSRCNN & DeepJF & CMSR\\
        \hline
            PSNR & U-VT & 26.789 & 27.988 & 27.968  & 27.036 & \textbf{29.928} \\
        \hline
            SSIM & U-VT & 0.8567 & 0.8202 & 0.8196 & 0.8410 & \textbf{0.882} \\
        \hline
            PSNR & SMB & 27.784 & 27.925 & 28.189 & 26.124 & \textbf{28.652} \\
        \hline
            SSIM  & SMB & 0.9140 & \textbf{0.9547} & 0.9386 & 0.8865 & 0.9341\\
        \hline
            PSNR & NIR & 28.807 & 30.665 & 30.143 & 27.094 & \textbf{31.201} \\
        \hline
            SSIM  & NIR & 0.8931 & 0.9005 & 0.8837 & 0.8694 & \textbf{0.9200} \\
        \hline
\end{tabular}

\captionof{table}{We compared CMSR to competing cross-modal SR methods, VTSRCNN and VTSRGAN (\cite{almasri2018multimodal}) and Deep Joint Filtering (\cite{DBLP:journals/corr/abs-1710-04200}), on the various datasets. (ULB17-VT, \textit{Shuffled}-Middlebury, EPFL NIR) and have taken the mean PSNR / SSIM scores, measured against the modality $4x$ GT versions.}
\label{table:avg_table1}
\end{table*}


\subparagraph{\textbf{Single Modality.}}

We evaluated CMSR against the baseline state-of-the art single modality method, ZSSR. (\cite{ DBLP:journals/corr/abs-1712-06087}) Our experiment shows that our method leverages the fine details in its RGB input and produces a SR outputs that are closer to a Ground-Truth version, as shown numerically in Table \ref{table:avg_table1} and visually in Figures \ref{fig:fence_comparison} and in the our NIR evaluation. (supplementary material Figure 1)

\subsection{Analysis}
\begin{figure}
\centering
\includegraphics[height=80pt,width=\linewidth]{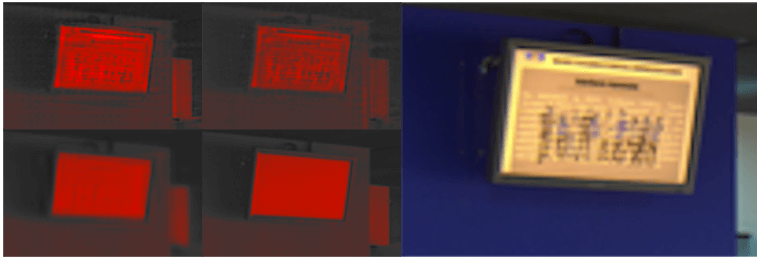}
\caption{
CMSR uses its RGB input conservatively. 
Compared to VTSRCNN (top left) and VTSRGAN (top right), CMSR avoids introducing noticeable redundant artifacts and textures induced by RGB modality. Ground-Truth (bottom right) is given as reference.}
\label{fig:artifact_comparison}
\end{figure}

\subparagraph{\textbf{RGB Artifacts.}} A fusion of multiple image sources, often causes the transfer of unnecessary artifacts. Those artifacts sabotage the image and harm its characteristics. Our method learns only the relevant RGB information that improves SR results; Figures \ref{fig:maagad_deformation_comparison}, \ref{fig:artifacts_rgbnir}, \ref{fig:vase_comparison} and \ref{fig:artifact_comparison} show cases where the RGB input contains a great amount of textural information, yet our SR output remains texture-free. 

\begin{figure}
\includegraphics[width=\linewidth]{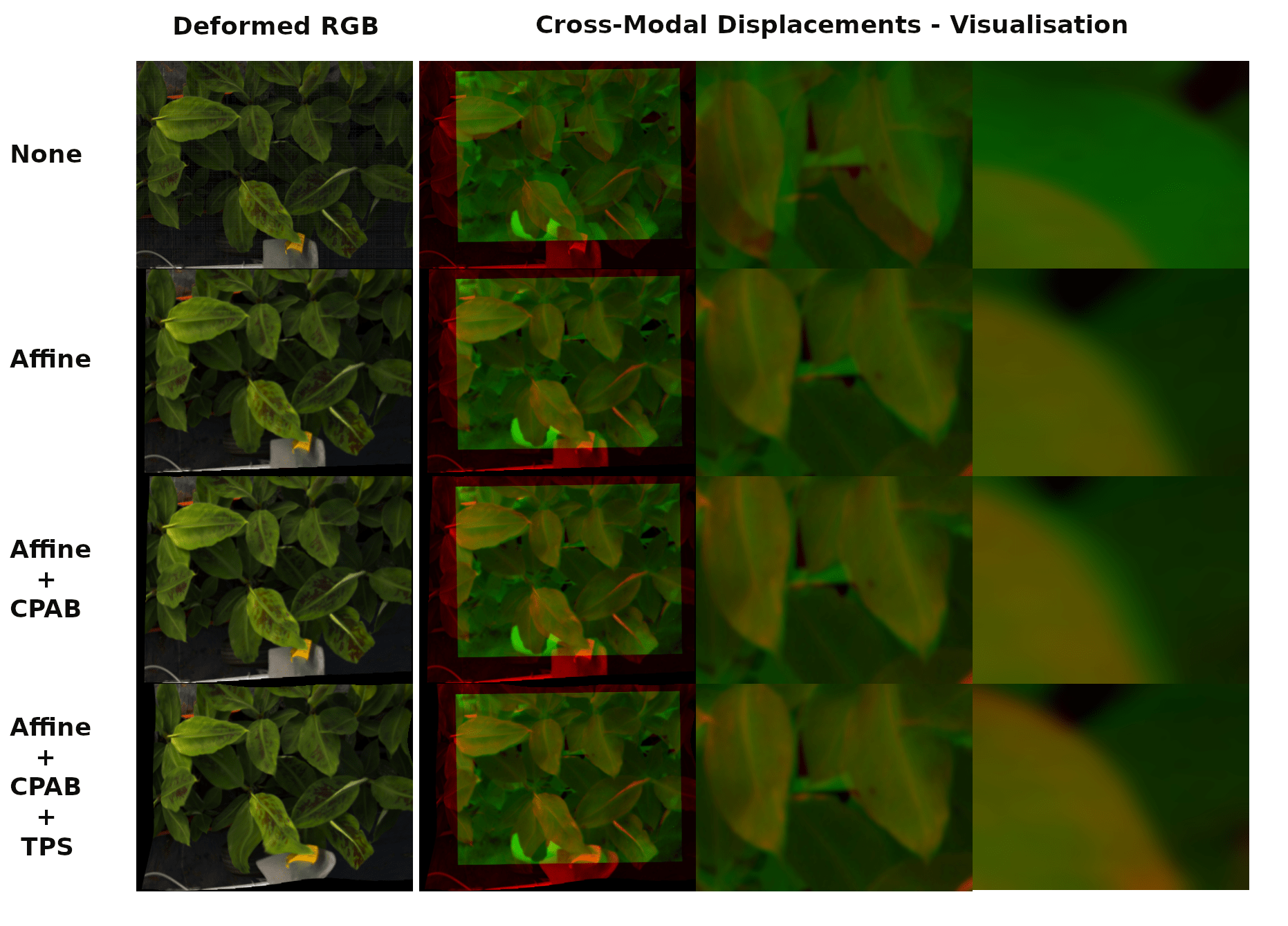}
\caption{We evaluated CMSR using different transformation layers. In the leftmost column, the resulting deformed RGB image is given. In the other columns we show the resulting alignment, visualized through blending of the R-G (Red-Green) channels of the aforementioned deformed RGB image, together with the Ground-Truth thermal image.}
\label{fig:ablation_deformation_visualisation}
\end{figure}

\begin{figure}
\centering
\includegraphics[width=235pt]{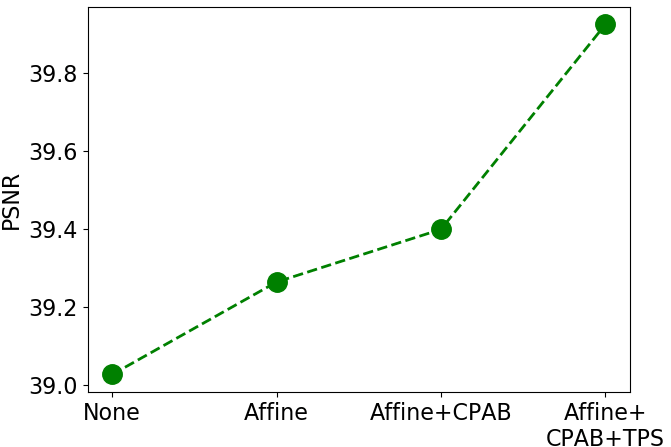}
\caption{We let CMSR perform $4x$ SR on a Weakly Aligned visual-thermal pair, with different transformation layers, averaged across 5 runs. The results indicate that each layer contributes to the final PSNR, which can also be seen visually in Figure \ref{fig:ablation_deformation_visualisation}}. 
\label{fig:ablation_deformation_graph}
\end{figure}

\subparagraph{\textbf{Local Deformation Ablation.}} As shown in Figure \ref{fig:deformation_vase}, our method deforms the RGB modality, for better alignment, without using any aligned RGB ground-truth images. Our deformation component provides the network the ability to align only details that assist and adhere to the super-resolution task, rather than committing to an image-to-image alignment per se.

To show the necessity of each layer of our coarse-to-fine deformation framework, we evaluated CMSR on a weakly aligned pair, adding one layer at a time and averaging across multiple runs. The results indicate that each layer is necessary and plays a different role in the alignment process as can be seen visually in Figure \ref{fig:ablation_deformation_visualisation}, and numerically in the supplementary material.
Our goal is not to perform perfect registration between the images, but rather to align only the necessary details to improve the quality of the higher resolution output. Hence, we measure the quality of the alignment through the generated SR result (e.g. in Figures \ref{fig:maagad_deformation_comparison} and \ref{fig:vase_comparison}), and not by conventional image registration metrics.




\section{Conclusions}
    We have introduced CMSR, a method for cross-modality super-resolution.

\textbf{Novelty} CMSR presents a novel way to tackle cross-modality SR; it achieves state-of-the-art results, qualitatively (visually) and quantitatively, using a minimalistic, easy to implement architecture, applied directly to any modality pair without pretraining. 

\textbf{Single Pair.} As a \textit{self-supervised} method, CMSR no training data, a prominent advantage when dealing with scarce and unique modalities. It adapts to the specifics of the given input pair, including among others: (i) the specific cross-modal misalignment that exists within the input pair and (ii) the degree and the manner in which the guiding modality should be incorporated. 

\textbf{Misalignment.}
A unique property of our method is that it is robust to cross-modal misalignment. This property is imperative, since in real life conditions, sight misalignment is, more often than not, unavoidable.
It should be emphasized that the alignment is done without pre-training or any supervision.


In the future, instead of deforming the entire RGB image once, we would like to deform different RGB objects differently, possibly using semantic segmentation, for further enhancement.

\twocolumn[
\begin{@twocolumnfalse}
\begin{center}
\textbf{\Large Weakly Aligned Joint Cross-Modality Super Resolution -\\}
\textbf{\Large Supplementary Material\\}
\end{center}
\end{@twocolumnfalse}
]

\section{Additional Results}
\begin{figure*}
\centering
\hspace{-5pt} ZSSR \cite{ DBLP:journals/corr/abs-1712-06087} \vspace{7pt} \hspace{20pt} DeepJF \cite{DBLP:journals/corr/abs-1710-04200} \hspace{10pt} VTSRCNN \cite{almasri2018multimodal} \hspace{5pt} VTSRGAN \cite{almasri2018multimodal} \hspace{20pt} CMSR \hspace{50pt} GT \hspace{35pt} RGB Input 
\makebox[\textwidth][c]{\includegraphics[height=550pt,width=\linewidth]{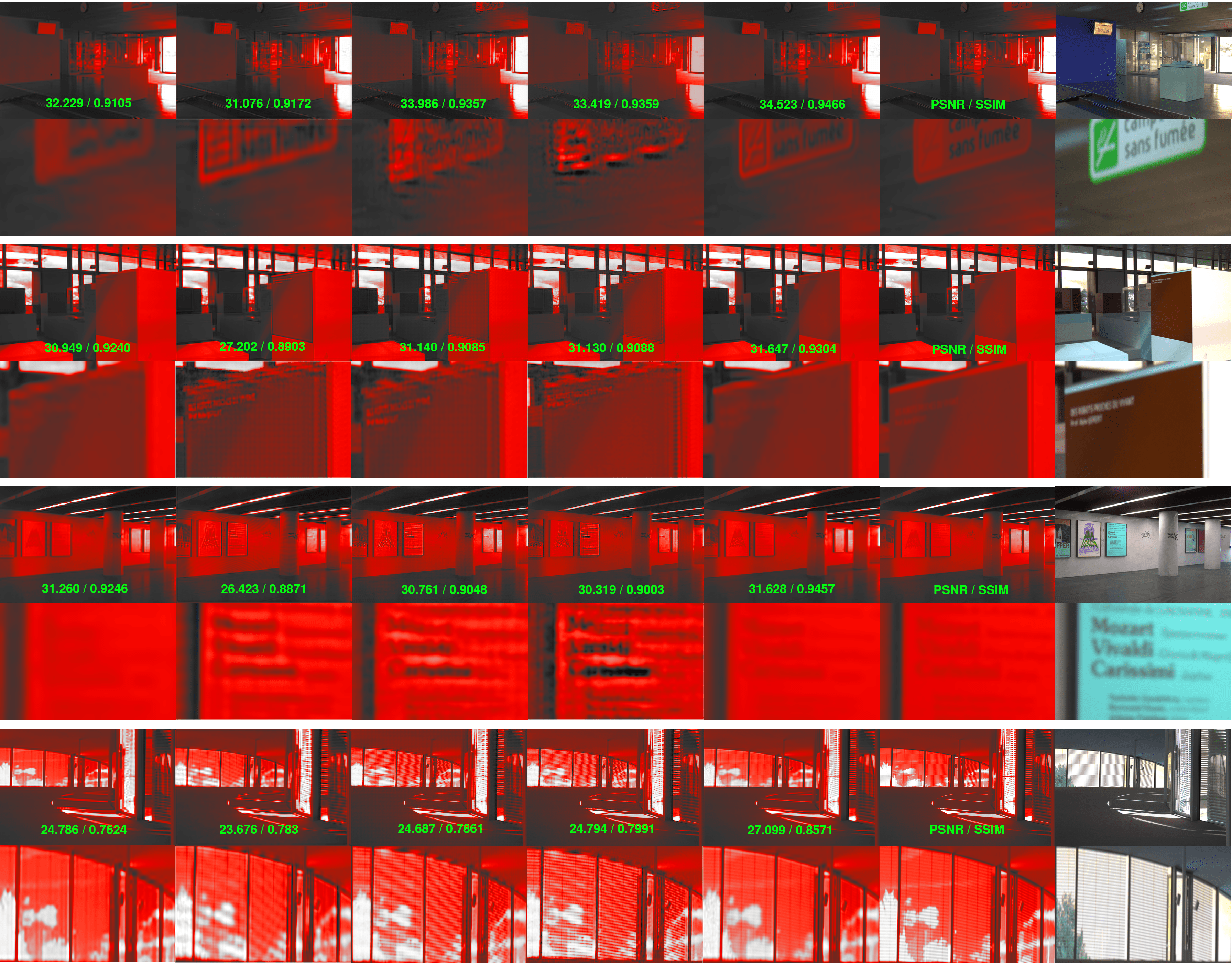}}
\caption{
We compared CMSR both to its single-modality baseline, ZSSR, and to competing cross-modality methods VTSRCNN, VTSRGAN and Deep Joint Filtering, on the NIR modality, in the task of $x \ 4$ SR. Our method, CMSR, is able to produce better super-resolved images visually and numerically, despite not being previously trained.}
\label{fig:nir}
\end{figure*}

    In Figure \ref{fig:nir}, additional results from our evaluation on the EPFL NIR dataset \cite{ivrl} are included. CMSR surpasses
state-of-the-art cross-modal methods, despite the fact that those competing methods were pre-trained extensively on the full dataset.
    
\section{Alternating Scales}

In \textbf{Section 3.2} of the submitted paper, the \textit{Alternating Scales} technique is briefly discussed. It corresponds to training CMSR using two different scales, alternating between them across iterations. Here, we wish to further elaborate on this technique.

\subsection{Alternating Scales - Elaboration}
Denoting our desired SR ratio (e.g. $2x$, $4x$) by $r$, our network, CMSR, takes a target modality input of size $H \ x \ W$ alongside with an RGB input of size $rH \ x \ rW$, and produces a target modality output of size $rH \ x \ rW$. Hence, by design, a ratio of $r$ must be preserved between CMSR's two inputs (The architecture of CMSR is given in the original paper, Figure 6).
Since CMSR is trained to reconstruct a random patch taken from its modality input (Figure 4 of the original paper, {\em{Training process}}), this random patch is down-sampled, by ratio $r$, before it is reconstructed by the CMSR network.
However, since the ratio between CMSR's two inputs must remain $r$, the corresponding RGB patch is also down-sampled accordingly, by ratio $r$. This way, we preserve the same ratio between CMSR's two input patches, as needed.

Nonetheless, instead of down-sampling the RGB patch to match this required ratio, it is also possible to naively up-sample the modality patch by ratio $r$. Clearly, this has the same effect on the ratio between the two patches, which yet again remains $r$. However, this way, we obtain a different training scheme. Figure \ref{fig:orig_vs_alternative} compares the two different schemes, corresponding to the two different scales CMSR operates on.

We found that by alternating between the two schemes during training, we are able to significantly improve our results. We name this combination of training schemes as the \textit{Alternating Scales} technique. It allows our network to be optimized using patches of their original scale, as explained in Table \ref{table:scales}. We observe that training our network on patches of their \textbf{original} scale improves its generalization capabilities, since during the inference stage, the network operates on the full input pair, at its \textbf{original} scale.

\subsection{Alternating Scales - Ablation Study}
\begin{figure*}
\centering
\includegraphics[height=170pt]{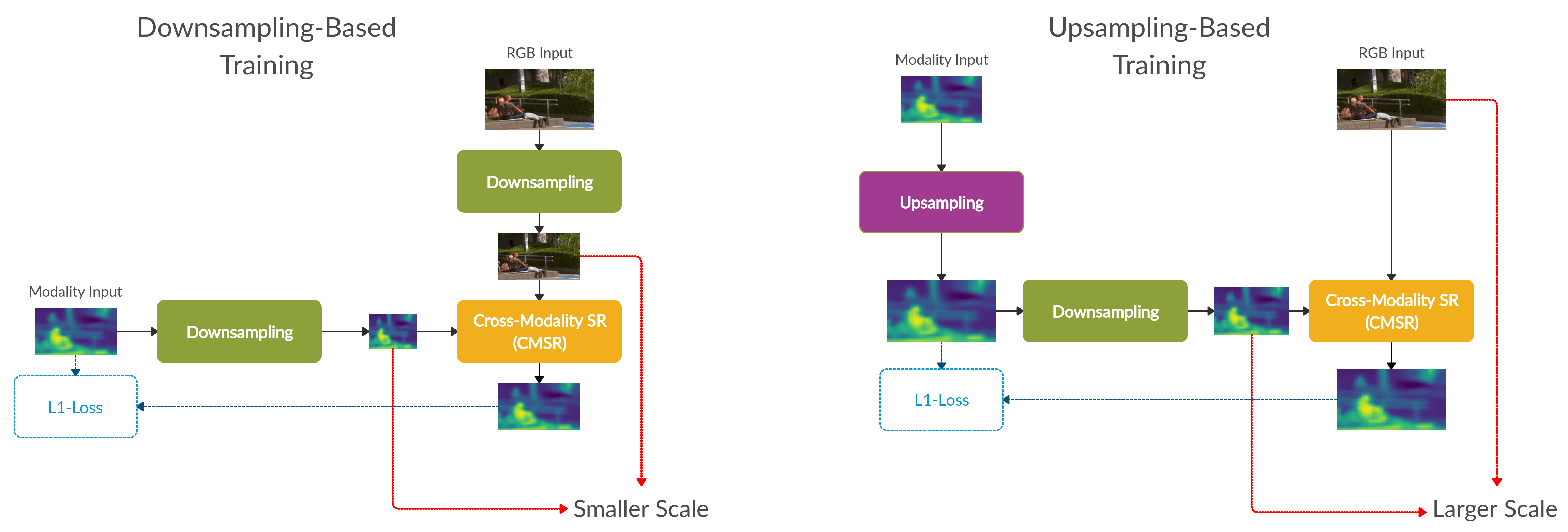}
\caption{The difference between the two training schemes lies in the scale CMSR (in Orange) operates on. The two schemes start with the exact same input pair, but in the Upsampling-Based training scheme (right), CMSR is fed inputs of larger scale. This scale difference is also explained in Table \ref{table:scales}.}
\label{fig:orig_vs_alternative}
\end{figure*}

We have conducted an experiment to show the improvement obtained by the \textit{Alternating Scale} technique. We trained CMSR using the two schemes (see Figure \ref{fig:orig_vs_alternative} and Table \ref{table:scales} for information on the schemes), alternating between them randomly. We used the Upsampling-Based scheme with probability $p$ and the Downsampling-Based with probability $1-p$.

\begin{figure}
\includegraphics[height=130pt,width=\linewidth]{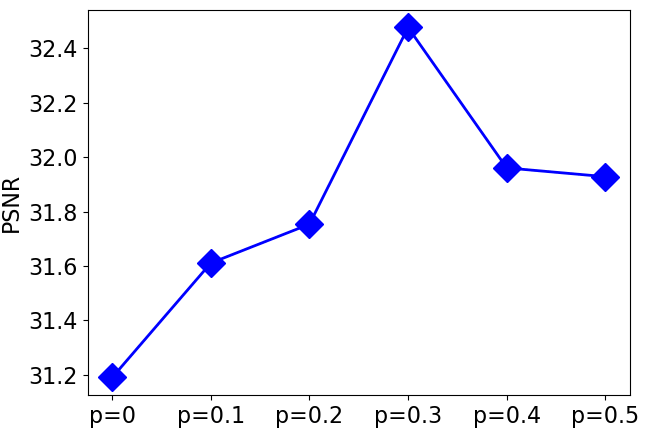}
\caption{We evaluated CMSR using different alternation probabilities. Namely, we trained it using the Upsampling-Based training scheme (Figure \ref{fig:orig_vs_alternative}) in fraction $p$ iterations, and using the Downsampling-Based scheme in the remaining fraction $1-p$. We averaged this experiment across multiple runs. According to the results, $p=0.3$ yields the best PSNR (32.476 $dB$). This can be also seen visually, in Figure \ref{fig:ablation_alternating_crops}}
\label{fig:ablation_alternating_graph}
\end{figure}

\begin{figure}
\centering
\includegraphics[height=95pt]{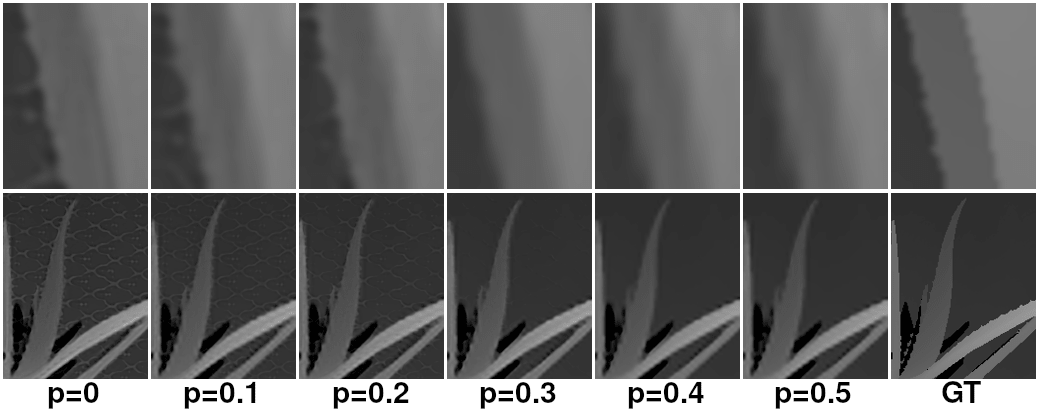}
\caption{We compare two patches taken from the Alternating Scales ablation study results, summarized in Figure \ref{fig:ablation_alternating_graph}. According to our experiment, the best SR result is obtained when using $p=0.3$ as the alternation probability.}
\label{fig:ablation_alternating_crops}
\end{figure}

According to the results, summarized in Figures \ref{fig:ablation_alternating_graph} and \ref{fig:ablation_alternating_crops}, the best PSNR was obtained when $p=0.3$, which starts decaying when $p>0.3$. We notice that $p>0$ always yields better results than $p=0$. This observation is important, since the risk of using sub-optimal $p$ values on new, unseen input pairs is minimal; using this technique is always better than not using it, regardless of $p$.

\begin{table}
\begin{centering}
\begin{tabular}{|c|c|c|}
        \hline
            Training Scheme & Modality Scale & RGB Scale \\
        \hline
            Down-sampling & \textbf{Original} & Down-scaled  \\
        \hline
            Up-sampling & Up-scaled & \textbf{Original}  \\
        \hline
\end{tabular}
\caption{In the Downsampling-Based training scheme, CMSR takes a down-sampled RGB input patch, but its modality input patch is reconstructed at its true, original scale. However, in the Upsampling-Based scheme, CMSR takes an original RGB input patch, at its true scale, but reconstructs a modality patch that was up-sampled beforehand.}
\label{table:scales}
\end{centering}
\end{table}

\clearpage
    
\section{Acknowledgements}
    This research was supported by a generic RD program of the Israeli innovation authority, and the Phenomics consortium. 
    
{\small
\bibliographystyle{ieeetr}
\bibliography{main}
}

\end{document}